\documentclass[journal abbreviation]{copernicus}

\usepackage{amsmath,amssymb}
\usepackage{dsfont}
\usepackage{graphicx}
\usepackage{subcaption}
\usepackage{cuted}
\usepackage{booktabs}
\usepackage{microtype}
\usepackage{hyperref}
\usepackage{natbib}

\usepackage{siunitx}
\usepackage{xcolor}

\title{IRENE: A Convolutional GRU Ensemble Model for Radar Precipitation Nowcasting over Italy}

\Author[1,*]{Alessandro}{Camilletti}
\Author[1,*]{Gabriele}{Franch}
\Author[1]{Elena}{Tomasi}
\Author[1]{Marco}{Cristoforetti}

\affil[1]{Fondazione Bruno Kessler}
\affil[*]{These authors contributed equally to this work.}

\correspondence{Alessandro Camilletti (acamilletti@fbk.eu)}

\runningtitle{IRENE -- Italian Radar Ensemble Nowcasting Experiment}

\runningauthor{Camilletti \& Franch}

\received{}
\pubdiscuss{}
\revised{}
\accepted{}
\published{}

\begin{document}

\firstpage{1}

\maketitle

\nolinenumbers

\begin{abstract}
We present IRENE (Italian Radar Ensemble Nowcasting Experiment), a deep learning model for probabilistic short-range precipitation
nowcasting over the Italian domain at \SI{1}{km} spatial and \SI{5}{min} temporal
resolution. IRENE adopts an encoder--forecaster architecture built on multi-scale
Convolutional Gated Recurrent Units (ConvGRUs), trained on the national radar
composite produced by the Italian Civil Protection Department (DPC). An
importance-sampling scheme focuses training on precipitation-relevant events, while
the almost-fair Continuous Ranked Probability Score (afCRPS) is adopted as the
primary probabilistic loss function. Two additional training configurations are proposed: an adversarial (GAN)
variant, IRENE-GAN, designed to improve the spatial sharpness of the generated
forecasts, and a spectrally constrained variant, IRENE-GAN-RAPSD, in which the
adversarial objective is complemented by an explicit penalty on the radially
averaged power spectral density. The three configurations are evaluated against
the stochastic extrapolation method STEPS and the pre-trained deep learning
model DGMR. All IRENE configurations attain a lower Continuous Ranked
Probability Score than both benchmarks at every lead time and rank
histograms closer to uniformity, indicating better probabilistic skill and ensemble calibration. In
terms of ensemble-mean mean absolute error the advantage is confined to the
first \SI{90}{min}, beyond which the strongly damped DGMR fields and, to a
lesser extent, STEPS become competitive. Spectral analysis shows that the
adversarial training removes the progressive loss of small-scale variance
exhibited by IRENE, at the cost of an excess of fine-scale power at long lead
times that the spectral penalty only partially controls.
\end{abstract}

\introduction[Introduction]

High-resolution precipitation nowcasting, with lead times ranging from a few minutes to about two hours, is a key component of modern hydrometeorological early-warning systems. It is particularly important for convective storms, urban hydrology, agriculture, and other weather-sensitive sectors where rapid changes in rainfall can have substantial societal and economic impacts.
Operational stakeholders require short-range forecasts with both high spatial resolution and reliable uncertainty quantification to support civil protection, hydrological modeling, and critical infrastructure management. 

Radar-based precipitation nowcasting has a long history, with early operational systems relying on optical-flow-based extrapolation of radar reflectivity fields. Classical Lagrangian extrapolation and stochastic ensemble schemes such as the Short-Term Ensemble Prediction System (STEPS) \citep{bowler2006} combine Lagrangian advection and stochastic perturbations to provide probabilistic nowcasts. However, these methods struggle in strongly non-stationary convective regimes, where optical-flow techniques are not sufficient. 

More recently, deep learning has been applied to nowcasting, with convolutional recurrent models such as ConvLSTM and TrajGRU, as well as encoder--decoder architectures operating on radar mosaics producing deterministic forecasts~\citep{shi2015,shi2017,agrawal2019,ayzel2020,franch2020}. Deep generative models have further advanced the state of the art by explicitly modelling the distribution of future radar fields. Notable examples include DGMR~\citep{ravuri2021}, which uses a GAN-based formulation to generate sharp and realistic ensemble forecasts, and GPTCast~\citep{franch2025}, which tokenizes radar fields and applies a GPT-style forecaster to model precipitation evolution probabilistically. Diffusion models have also been proposed for high-resolution precipitation nowcasting~\citep{asperti2024}. 
Yet, none of the existing works are trained on the entire Italian radar system, limiting their direct applicability to the Italian domain. IRENE is designed to bridge this gap by providing a robust probabilistic precipitation nowcasting model trained directly on the Italian radar archive. 

In Italy, the national radar composite maintained by the Civil Protection Department (DPC) provides \SI{1}{km}, \SI{5}{min} surface rainfall estimates over the entire country, making it a natural target for nowcasting applications. In this work, we use the Analysis-Ready and Cloud-Optimised (ARCO) archive of this radar product introduced in \citet{franch2026} to train and evaluate IRENE, a multi-scale encoder--forecaster model based on Convolutional Gated Recurrent Units (ConvGRUs). The underlying architecture was originally presented in \cite{shi2015,shi2017} and developed for deterministic nowcasting; here it is extended to probabilistic forecasting by injecting stochastic noise at the coarsest decoder scale to generate an ensemble of trajectories (Sect.~\ref{sec:stochastic_ensemble_head}) and by training with the almost-fair CRPS (afCRPS) as the primary objective (Sect.~\ref{sec:train_ensemble}). Training sequences are selected from the whole archive by an importance sampler that favours relevant precipitation events. The almost-fair Continuous Ranked Probability Score (afCRPS) is used to guide the training, in order to obtain ensembles that are both consistent with the observed precipitation and whose spread correctly represents the forecast uncertainty.

The present paper focuses on the model architecture, the training configuration, the importance-sampling algorithm used to select relevant sequences, and the evaluation methodology of the trained model, while the Analysis-Ready and Cloud-Optimised (ARCO) archive of the Italian radar product is described in detail in \citet{franch2026}.

\section{IRENE Model Architecture}
\label{sec:architecture}

Let $x_{t-n+1}, \dots, x_t$ denote a sequence of $n$ radar composite images, each defined on a regular cartesian grid of \SI{1}{km} resolution and \SI{5}{min} temporal sampling.
The goal is to predict the subsequent $m$ radar images $x_{t+1}, \dots, x_{t+m}$ on the same grid as an ensemble of forecasts.
Formally, the model learns a mapping from past fields to a predictive distribution over future fields, $(x_{t-n+1}, \dots, x_t) \mapsto \mathcal{P}(x_{t+1}, \dots, x_{t+m})$.

IRENE's architecture is divided into a multi-scale ConvGRU encoder and an autoregressive ConvGRU forecaster, where the encoder produces a latent representation that the forecaster uses to generate the ensemble forecast.

\subsection{Multi-scale ConvGRU encoder}
\label{sec:encoder}

IRENE employs a multi-scale Convolutional Gated Recurrent Unit (ConvGRU) encoder to extract spatio-temporal features from the input radar sequence.
The encoder is composed of a hierarchy of ConvGRU blocks arranged from fine to coarse spatial scales. At each level, a ConvGRU processes the full input sequence along the temporal dimension and updates its hidden state at every time step. After the recurrent update, the feature maps are downsampled by a factor of two in each spatial dimension using a PixelUnshuffle operation, which trades spatial resolution for channel depth. Repeating this pattern across successive blocks produces a multi-scale representation in which temporal information is encoded jointly with increasingly coarse spatial context.

Let $h^s_t$ denote the hidden state at scale $s$ and time $t$, and let $x_t^s$ be the corresponding input feature map.
The ConvGRU dynamics at scale $s$ are given by
\begin{align}
  Z_t^s &= \sigma\left(W_{xz}^s * x_t^s + W_{hz}^s * h_{t-1}^s + b_z^s\right), \\
  R_t^s &= \sigma\left(W_{xr}^s * x_t^s + W_{hr}^s * h_{t-1}^s + b_r^s\right), \\
  \tilde{h}_t^s &= \tanh \left( W_{x\tilde{h}}^s * x_t^s + W_{h\tilde{h}}^s * \left(R_t^s \odot h_{t-1}^s\right) + b_{\tilde{h}}^s \right), \\
  h_t^s &= \left(1 - Z_t^s\right) \odot h_{t-1}^s + Z_t^s \odot \tilde{h}_t^s,
\end{align}
where $*$ denotes convolution, $\odot$ denotes element-wise multiplication, and $\sigma$ is the logistic sigmoid function.
The update, $Z_t^s$, and reset, $R_t^s$, gates control how much information is retained from the previous hidden state and how much new information is incorporated from the current input.
After processing all $n$ input frames, the encoder outputs the final hidden states $\{h^{(s)}_t\}_s$ at all scales, which summarise the past precipitation evolution.

\begin{figure*}
  \centering
  \includegraphics{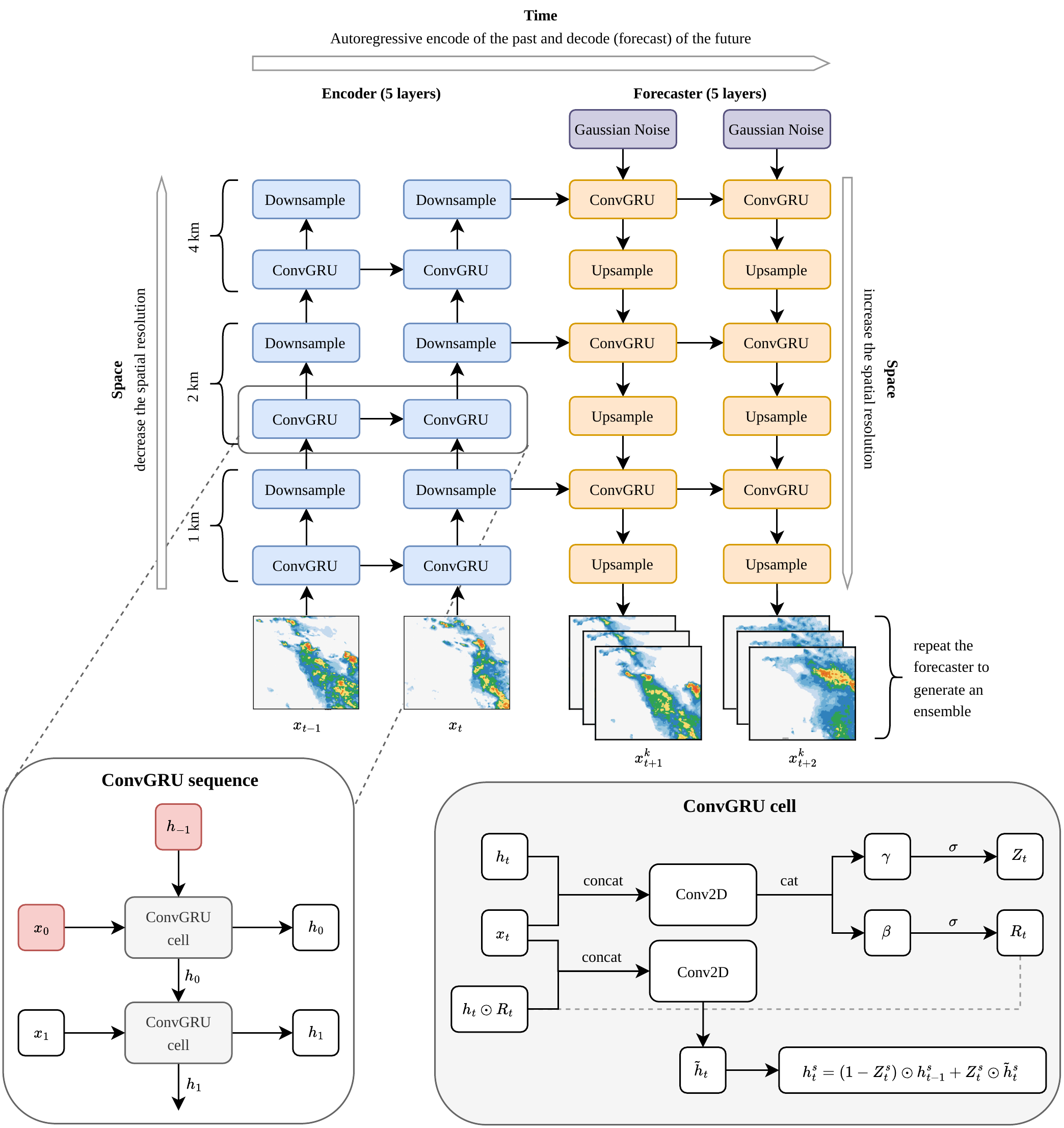}
  \caption{Schematic representation of the IRENE architecture.
  The encoder (top left) processes the past radar sequence through a multi-scale ConvGRU hierarchy, from \SI{1}{km} to coarser resolutions.
  The forecaster (top right) autoregressively generates future frames using ConvGRU layers and upsampling operations, optionally conditioned on stochastic noise for ensemble generation.
  In the bottom, the ConvGRU sequence and cell are described}
  \label{fig:IRENE_architecture}
\end{figure*}

In the configuration used throughout this work the hierarchy comprises $S = 5$
blocks, so that the representation is progressively coarsened from \SI{1}{km} to
\SI{32}{km}. 
The number of channels quadruples at every downsampling (PixelUnshuffle operation) step and the ConvGRU hidden size at each scale equals its input size; the channel widths are therefore not free parameters but follow from the input depth (Table~\ref{tab:architecture}).
All gates use $3 \times 3$ convolutions, and the update and reset gates are
computed by a single convolution whose output is split in two. For clarity,
Fig.~\ref{fig:IRENE_architecture} sketches only three of the five scales. The
resulting encoder has $\approx 3.8 \times 10^{6}$ trainable parameters.

% Figure: architecture schematic

\subsection{Autoregressive ConvGRU forecaster}
\label{sec:forecaster}

The forecaster mirrors the encoder with a multi-scale ConvGRU hierarchy that operates from coarse to fine resolution and generates future radar frames in a recurrent autoregressive rollout. At the initial forecast time step, the forecaster is seeded with the encoder hidden states at each scale. In ensemble mode, stochastic noise is injected as the input to the decoder ConvGRU cells, introducing variability across ensemble members (see Sec. \ref{sec:stochastic_ensemble_head}).

For each lead time $\tau = 1, \dots, m$, the forecaster updates the ConvGRU hidden states and produces a set of intermediate feature maps at each scale. PixelShuffle layers upsample and propagate information from coarse to fine scales, each of them dividing the number of channels by four; the last block therefore returns a single-channel field at \SI{1}{km}, and no additional projection head is required. The predicted field is finally clipped to the physical range of the transformed variable defined in Sect.~\ref{sec:preprocessing}. The forecaster has $6.0 \times 10^{7}$ trainable parameters, an order of magnitude more than the encoder, since the coarsest and widest block (\SI{32}{km}, 1024 channels) is evaluated first.

The forecast is therefore generated autoregressively in hidden-state space: each step depends on the previous recurrent state, while the previous predicted radar field is not explicitly fed back as an input.

\subsection{Stochastic ensemble head}
\label{sec:stochastic_ensemble_head}

To generate probabilistic forecasts, the forecaster is driven by stochastic
noise rather than by a deterministic input. The coarsest decoder block receives
no feature input from a previous block; the input is, at every lead
time, a Gaussian noise $\eta^{(k)}_\tau \sim \mathcal{N}(0, I)$ of the
same shape as the coarsest hidden state, drawn independently for each ensemble
member $k = 1, \dots, K$ and each $\tau$. The finer decoder blocks are then fed
deterministically with the upsampled output of the block above, so that the
stochasticity introduced at the \SI{32}{km} scale is propagated and refined
towards \SI{1}{km} through the recurrent hidden states. Resampling the noise at
every lead time, rather than drawing a single latent vector per member, lets the
members diverge progressively as the forecast evolves, which is the behaviour
required for the spread to grow with lead time.
The forecaster is then run forward for $m$ time steps, producing a stochastic trajectory $\{\hat{x}^{(k)}_{t+\tau}\}_{\tau=1}^m$.

Repeating this procedure for $K$ ensemble members yields an empirical predictive distribution at each grid point and lead time.
The ensemble size $K$ can be chosen at training time for the loss evaluation and can be increased at inference time if computational resources allow. Together, the encoder and forecaster form a generator of $6.4 \times 10^{7}$ trainable parameters. Unless stated otherwise, all IRENE results in this paper use $K = 10$ at training and inference.

The resulting configuration is summarised in Table~\ref{tab:architecture}.

\begin{table}[t]
  \caption{Configuration of the IRENE encoder--forecaster, common to the three
  training configurations. The
  $S = 5$ blocks connect $S+1 = 6$ resolution levels.}
  \label{tab:architecture}
  \begin{tabular}{ll}
    \toprule
    Input sequence & 6 frames (\SI{30}{min}) \\
    Training forecast horizon & 12 frames (\SI{60}{min}) \\
    Verification forecast horizon & 24 frames (\SI{120}{min}) \\
    Patch size & $256 \times 256$ pixels (\SI{1}{km}) \\
    Number of blocks, $S$ & 5 \\
    Encoder resolutions ($S+1$ levels) & 1, 2, 4, 8, 16, \SI{32}{km} \\
    Encoder channels & 1, 4, 16, 64, 256, 1024 \\
    Forecaster resolutions ($S+1$ levels) & 32, 16, 8, 4, 2, \SI{1}{km} \\
    Forecaster channels & 1024, 256, 64, 16, 4, 1 \\
    \bottomrule
  \end{tabular}
\end{table}

\section{Dataset}
\label{sec:dataset}
 
\begin{figure}
  \centering
  \includegraphics{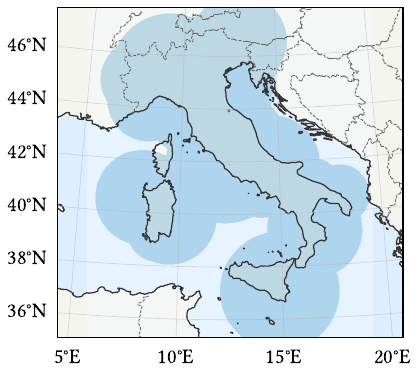}
  \caption{Spatial domain of the IT-DPC-SRI dataset. Shaded area indicates radar coverage, defined as grid cells containing at least one valid observation.}
  \label{fig:domain}
\end{figure}

The training and evaluation dataset is derived from the Italian IT-DPC-SRI radar composite, which provides a national mosaic of surface rainfall estimates at \SI{1}{km} spatial resolution and \SI{5}{min} temporal resolution.
A long-term archive of this composite, covering more than a decade of observations, is ingested and harmonised into a single Zarr-based ARCO datacube.
The harmonisation includes regridding to a common domain and projection, handling missing data and format changes, and applying consistent metadata.
An overview of the spatial coverage of the Italian domain is shown in Fig.~\ref{fig:domain}.
The details of the ARCO datacube construction are described in \citet{franch2026}. 
%An example of precipitation events is shown in Fig. \ref{fig:sample_precipitation}, representing the Emilia-Romagna flood event of 16–17 May 2023, characterized by intense convective activity.
%\begin{figure*}
%  \centering
%  \includegraphics{plots/sample_precipitation}
%  \caption{Precipitation rate (mm h$\mathrm{^-1}$) during the Emilia-Romagna flood event of 16–17 May 2023. The six panels, spaced 3 hours apart, illustrate the temporal evolution of the rainfall system.}
%  \label{fig:sample_precipitation}
%\end{figure*}
 
For the IRENE experiments, the period from 1 January 2021 to 11 December 2025 is
used for model training, validation, and testing.
We train the model on $256\times 256$ patches of $18$ timesteps, with six provided as input and twelve
as target ground truth. Training sequences are drawn from the archive in two
stages: a data-cleaning step (Sect.~\ref{sec:data_cleaning}) that filters
candidate space--time cubes for data quality, followed by an importance-sampling
step (Sect.~\ref{sec:importance_sampling}) that selects meteorologically
relevant sequences from the cleaned pool.
 
\subsection{Preprocessing}
\label{sec:preprocessing}

The models do not operate on rain rates directly. Each rain-rate field $R$
(mm\,h$^{-1}$) is first converted to equivalent reflectivity through the
Marshall--Palmer relation $Z = 200\,R^{1.6}$, expressed in decibels,
\begin{equation}
  \mathcal{Z} = 10 \log_{10}\left(200\,R^{1.6}\right),
  \label{eq:marshall_palmer}
\end{equation}
clipped to the interval $[0,60]$\,dBZ and linearly rescaled to $[-1, 1]$. The
inverse transformation is applied to the model output before verification, so
that all scores reported in Sect.~\ref{sec:results} are expressed in
mm\,h$^{-1}$. This choice compresses the strongly skewed rain-rate distribution
into a bounded variable with an approximately uniform dynamic range, which is
better conditioned for optimisation. It also imposes two physical bounds on the
forecasts: rain rates below $0.037$\,mm\,h$^{-1}$, corresponding to $0$\,dBZ, are mapped to the lower limit, and intensities are capped at $60$\,dBZ, i.e. approximately $205$\,mm\,h$^{-1}$. Both losses and
verification of the deep learning models are therefore insensitive to
variability outside this range. Note that the afCRPS loss (Sec.~\ref{sec:train_ensemble}) is computed in the transformed
space, so that a given error is penalised similarly whether it occurs at a low or a high rain rate, rather than errors at high rain rates being penalised more heavily, as they would be if the loss were computed directly in mm\,h$^{-1}$.
 
\subsection{Data Cleaning}
\label{sec:data_cleaning}
 
Candidate space--time cubes are enumerated on a regular grid covering the whole
archive and the full study period, with a stride of $3$ timesteps
(\SI{15}{min}) in time and of $16$ pixels (\SI{16}{km}) in each horizontal
direction; each candidate has the full $24 \times 256 \times 256$ cube
dimensions. Every candidate is screened against two data-quality criteria
before being made available to the importance sampler of
Sect.~\ref{sec:importance_sampling}.
The first criterion is temporal continuity: all $24$ radar acquisitions of a
candidate cube must be spaced by the nominal \SI{5}{min} interval, with no
missing timestep in the underlying archive.
The second criterion caps the number of missing (NaN) pixels within the cube at
$10^{4}$, i.e. less than $1\%$ of its $24 \times 256 \times 256 \approx
1.57 \times 10^{6}$ points.
Because this threshold constrains but does not eliminate missing data, up to
$10^{4}$ NaN pixels can remain in an accepted cube; these residual gaps are
replaced with the minimum value of the normalised input range
(Sect.~\ref{sec:preprocessing}), equivalent to the no-precipitation state,
before the sequence is passed to the network. Because the $1\%$ threshold is
applied to the full $256\times256$ cube, it preferentially retains candidates
drawn from the well-covered interior of the domain (Fig.~\ref{fig:domain}) and
excludes most candidates straddling the radar-coverage boundary, so the
training population under-represents precipitation behaviour near the edges
of the operational domain.
 
\subsection{Importance Sampling}
\label{sec:importance_sampling}
 
To focus training on meteorologically relevant precipitation events, we employ an
importance sampling strategy adapted from \citet{ravuri2021}. Starting from the
cleaned candidate pool of Sect.~\ref{sec:data_cleaning}, a candidate patch is
accepted with probability
\begin{equation}
  q = \min\left(1, q_{\min} + c \cdot \overline{\mathcal{R}}\right),
  \label{eq:importance_prob}
\end{equation}
where $\overline{\mathcal{R}}$ denotes the time- and space-averaged rain rate
within the patch, transformed via $\overline{\mathcal{R}} \mapsto 1 - e^{-\overline{\mathcal{R}}/s}$.
The parameters are set to $q_{\min} = 10^{-4}$, $c = 0.1$, and $s = 1$
mm\,h$^{-1}$, yielding a selection probability that increases monotonically with
average rainfall intensity while guaranteeing that even ``dry'' patches have a
small but non-zero chance of selection.
 
The retained cubes are split sequentially
into training (90\%), validation (5\%), and test (5\%) sets. The procedure retains $359\,000$ space--time cubes, which the sequential split
assigns to $323\,100$ training, $17\,950$ validation and $17\,950$ test samples.
%All verification results reported in
%Sect.~\ref{sec:results} are computed on the test subset, which spans from
%2025-07-28T02:25 to 2025-11-18T15:10 UTC (\SI{113}{days}); the validation
%subset immediately precedes it, from 2025-04-16T21:10 to 2025-07-28T02:20 UTC,
%and the training subset covers the remainder of the archive up to that point.
 
Because the candidate grid of Sect.~\ref{sec:data_cleaning} has a stride of
three timesteps while each cube spans twenty-four, consecutive samples can
share frames if their starting indices differ by less than the cube length.
This occurs exactly
at the train-validation and validation-test split boundaries. The leakage affects two samples out of $359\,000$ and cannot
influence the verification statistics.

\subsection{Data Augmentation}

To increase the effective sample size and improve model generalisation,
on-the-fly spatial and temporal data augmentation is applied during training.

Spatially, each sampled radar sequence is subjected to random transformations
that preserve the statistical structure of precipitation fields, specifically
90, 180 or 270-degree rotations and horizontal or vertical flips. Formally,
for each training sample, a random element $g$ is drawn from the dihedral group
$D_4$ of square symmetries and applied uniformly to both the input and target
frames. Composing these discrete rotations and reflections yields up to eight
distinct spatial augmentations per original sequence.

Temporally, we exploit the difference between the sampled patch length and the
required sequence length. While the importance sampler extracts continuous
datacubes of $24$ timesteps, the model requires only $18$ timesteps per sample
($6$ for the past context and $12$ for the forecast target). During loading,
the starting timestep $t_{\mathrm{start}}$ of the 18-step sequence is chosen
uniformly at random from the available 7-step sliding window within the
24-step patch. 

Combining the 8 spatial symmetries with the 7 possible temporal shifts yields
up to 56 distinct augmented views for every extracted radar patch.

\section{Training}
\label{sec:training}
The IRENE architecture can be trained under different paradigms depending on the
forecasting objective. We detail three configurations: a stochastic ensemble
formulation, a generative adversarial network (GAN) designed to produce
meteorologically realistic precipitation fields, and a spectrally constrained
variant of the latter, in which the adversarial objective is complemented by an
explicit penalty on the radially averaged power spectral density of the
forecast fields.

\subsection{Ensemble Configuration}
\label{sec:train_ensemble}

To quantify forecast uncertainty, IRENE is trained as a stochastic model that
produces an ensemble of future trajectories. At each training sample, $K = 10$
ensemble members are generated using independent noise realisations. The model
is optimised using a strictly proper scoring rule, specifically the almost-fair
Continuous Ranked Probability Score (afCRPS), that we describe below.

For a predictive cumulative distribution function (CDF) $F$ and an observed
value $y$, the CRPS is defined as
\begin{equation}
  \label{eq:crps_cdf}
  \mathrm{CRPS}(F, y) = \int_{-\infty}^{\infty} \bigl(F(z) - \mathds{1}_{\{z \ge y\}}\bigr)^2 \, \mathrm{d}z,  
\end{equation}
where $\mathds{1}_{\{z \ge y\}}$ is the indicator function. When $F$ is
represented by a finite ensemble $\{y^{(k)}\}_{k=1}^{K}$, the integral in
Eq.~\eqref{eq:crps_cdf} can be evaluated analytically, yielding the energy-form
estimator:
\begin{equation}
  \widehat{\mathrm{CRPS}}_K =
    \frac{1}{K} \sum_{k=1}^{K} \lvert y^{(k)} - y \rvert
    - \frac{1}{2K^2} \sum_{k=1}^{K} \sum_{j=1}^{K}
      \lvert y^{(k)} - y^{(j)} \rvert.
  \label{eq:crps_ensemble}
\end{equation}
Because the standard CRPS estimator is positively biased for finite $K$, we
follow \citet{lang2026} and employ the almost-fair CRPS (afCRPS) as the primary
training objective. For an ensemble forecast $\{y^{(k)}\}_{k=1}^{K}$ and
observation $y$, the $\alpha$-fair variant is defined as:
\begin{equation}
  \label{eq:afcrps_alpha}
  \mathcal{L}_{\mathrm{afCRPS}} =
  \frac{1}{2K(K-1)} \sum_{k=1}^{K} 
  \begin{aligned}[t]
    &\sum_{\substack{j=1 \\ j \neq k}}^{K}
    \Bigl(
      \lvert y^{(k)} - y \rvert
      + \lvert y^{(j)} - y \rvert \\
      &\qquad - (1-\varepsilon)\,\lvert y^{(k)} - y^{(j)} \rvert
    \Bigr),
  \end{aligned}
\end{equation}

where $\varepsilon = (1-\alpha)/K$ and $\alpha \in (0,1]$ controls the degree of
fairness. Setting $\alpha=1$ recovers the perfectly fair CRPS, while
$\alpha \lesssim 1$ slightly relaxes the spread term to avoid degeneracies for
small ensemble sizes. Following \citet{lang2026}, we use $\alpha = 0.95$.

To further regularize the temporal coherence of the ensemble trajectories and
prevent unrealistic frame-to-frame jumps, we introduce a temporal consistency
penalty:
\begin{equation}
  \label{eq:temporal_consistency}
  \mathcal{L}_{\mathrm{temp}} =
    \frac{1}{m-1} \sum_{\tau=1}^{m-1} \frac{1}{K} \sum_{k=1}^{K}
    \bigl\lvert y_{\tau+1}^{(k)} - y_{\tau}^{(k)} \bigr\rvert.
\end{equation}
The total reconstruction objective for the ensemble configuration is then given by:
\begin{equation}
  \label{eq:afcrps_total}
  \mathcal{L}_{\mathrm{rec}} =
    \mathcal{L}_{\mathrm{afCRPS}} +
    \lambda_{\mathrm{temp}} \,\mathcal{L}_{\mathrm{temp}},
\end{equation}
where $\lambda_{\mathrm{temp}} \geq 0$ is a tunable hyperparameter. This combined
loss encourages the model to accurately capture both the conditional mean and
the predictive spread while maintaining smooth temporal evolution. A value of 
$\lambda_{\mathrm{temp}} = 0.01$ was sufficient to guarantee temporal consistency between 
consecutive frames. 

\subsection{Generative Adversarial Network}
\label{sec:train_gan}

While the afCRPS loss produces well-calibrated probabilistic forecasts, models
optimised solely on pixel loss often struggle to generate the sharp,
high-frequency structural details characteristic of real precipitation fields.
To overcome this, we extend the ensemble configuration into a generative
adversarial network (GAN) framework.

In this setup, the ConvGRU encoder--forecaster acts as a generator $G$,
trained with the afCRPS-based reconstruction loss $\mathcal{L}_{\mathrm{rec}}$
(Eq.~\ref{eq:afcrps_total}), with $\lambda_{\mathrm{temp}} = 0$.
Simultaneously, a 3-D PatchGAN
discriminator network $D$ \citep{isola2017} is trained to classify overlapping
spatial patches as real (observations) or fake (model predictions). The
discriminator is a three-dimensional PatchGAN: three strided convolutional
blocks with $3 \times 4 \times 4$ kernels in the (time, latitude, longitude)
directions, batch normalisation and leaky ReLU activations, followed by two
unit-stride blocks that output a map of patch-wise logits. With the base width
of 16 used here (Table~\ref{tab:training}), this amounts to
$5.2 \times 10^{5}$ trainable parameters. Because the
convolutions are three-dimensional, the discriminator judges the joint
spatio-temporal structure of the forecast sequence rather than the texture of
isolated frames, and therefore also penalises temporally inconsistent
evolutions. It is trained with a hinge loss and, in IRENE-GAN, is activated only after a
warm-up of $4 \times 10^{5}$ optimisation steps, so that the
adversarial term starts acting on a generator that already produces
meaningful fields. By evaluating local patches rather than a single global score, the discriminator explicitly models the high-frequency structure, penalising locally unrealistic textures and encouraging the generator to produce sharp, fine-scale precipitation patterns.

The adversarial training employs a hinge loss to enforce a margin
between the discriminator outputs, thereby stabilising the training dynamics.
Let $x$ denote the true observed fields and $\hat{x}$ denote the model
predictions. The discriminator $D$ minimises the objective:
\begin{equation}
    \mathcal{L}_{D} = \mathbb{E}_{x} \left[ \max(0, 1 - D(x)) \right] + \mathbb{E}_{\hat{x}} \left[ \max(0, 1 + D(\hat{x})) \right].
\end{equation}
Conversely, the generator seeks to fool the discriminator by minimising the
adversarial loss:
\begin{equation}
    \mathcal{L}_{\text{adv}} = - \mathbb{E}_{\hat{x}} \left[ D(\hat{x}) \right].
\end{equation}
Importantly, at every training step the discriminator evaluates one ensemble
member drawn at random for each element of the batch, rather than the ensemble
mean, so that the adversarial penalty encourages realism in every distinct
stochastic realisation instead of in their average, which is by construction
smoother.

A common challenge in adversarial forecasting is balancing the adversarial
penalty with the primary reconstruction objective. If adversarial gradients
dominate, they can destabilise training and degrade overall forecast accuracy.
To mitigate this, we employ the dynamic adaptive weighting mechanism proposed
by \citet{esser2021}. At each training step, an adaptive weight $\lambda$ is
computed as the ratio of the gradient norms of the reconstruction and
adversarial losses, evaluated with respect to the weights of the generator's
final layer, $\theta_{\text{last}}$:
\begin{equation}
    \lambda = \frac{\lVert \nabla_{\theta_{\text{last}}} \mathcal{L}_{\text{rec}} \rVert}{\lVert \nabla_{\theta_{\text{last}}} \mathcal{L}_{\text{adv}} \rVert + \epsilon},
\end{equation}
where $\epsilon = 10^{-4}$ ensures numerical stability and $\lambda$ is clipped
to $[0, 10^{4}]$ and treated as a constant in the backward pass. The total objective for
the generator is finally formulated as:
\begin{equation}
    \mathcal{L}_{\text{total}} = \mathcal{L}_{\text{rec}} + w_D \lambda \mathcal{L}_{\text{adv}},
\end{equation}
where $w_D$ is a fixed scalar hyperparameter. This adaptive scaling ensures that
the adversarial gradients remain comparable with the reconstruction gradients
throughout training, allowing the model to smoothly refine high-frequency details
without compromising the probabilistic accuracy dictated by the afCRPS.
We set $w_D = 0.04$ in both adversarial configurations.

\subsection{Spectrally constrained GAN}
\label{sec:train_gan_rapsd}

The adversarial objective of Sect.~\ref{sec:train_gan} rewards locally
realistic texture, but it does not explicitly constrain how variance is
distributed across spatial scales. We therefore consider a third
configuration, denoted IRENE-GAN-RAPSD, in which the generator objective is
augmented with a direct penalty on the radially averaged power spectral
density (RAPSD) of the forecast fields.

The RAPSD is computed from a rain-rate field $r(x,y)$ defined on a grid of
size $N_x \times N_y$ with spacings $dx$ and $dy$. The field is first
multiplied by a two-dimensional Hann window $w(x,y)$ and transformed with a
two-dimensional discrete Fourier transform, giving the two-dimensional PSD
\begin{equation}
  P(k_x,k_y) =
    \lvert \hat r(k_x,k_y) \rvert^2 \, dx \, dy /
    \bigl( N_x N_y \langle w^2 \rangle \bigr),
  \label{eq:psd2d}
\end{equation}
where $k_x$ and $k_y$ are the Fourier wavenumbers and $\langle w^2 \rangle$ is
the mean squared window amplitude. Integration in circular coordinates yields
the radial spectrum
\begin{equation}
  \mathrm{PSD}_{\mathrm{rad}}(k) = k \int_{0}^{2\pi} P(k,\theta) \, d\theta,
  \label{eq:psd_rad}
\end{equation}
with $k = \sqrt{k_x^2 + k_y^2}$ the radial wavenumber and $P(k,\theta)$ the
two-dimensional PSD in polar coordinates. Equation~\eqref{eq:psd_rad}
describes how variance is distributed across spatial scales, from the domain
scale to the Nyquist limit, i.e. \SI{2}{km} for the radar data on which the
model is trained.

During training, Eq.~\eqref{eq:psd_rad} is estimated in a windowed,
Welch-like fashion rather than on the full field. Each predicted and observed
frame is decomposed into overlapping square windows of $64 \times 64$ pixels
with a stride of $32$ pixels ($50\%$ overlap), so that the constraint acts on
scales up to $\approx \SI{64}{km}$. Within each window the mean is subtracted
before tapering, so that errors in the local rain amount are left to the
reconstruction term instead of leaking across wavenumbers through the taper.
The spectrum is discretised into $n_b = 32$ radial wavenumber bins and
expressed in $\log_{10}$ units, and the window log-spectra are averaged over
window positions, yielding one log-spectrum per ensemble member and lead time.
Matching is therefore performed at the frame level rather than window by
window: pairing spectra by location would penalise members that legitimately
displace precipitation and would encourage the injection of spurious texture
into dry regions. Denoting by $\widehat{P}^{(k)}(k_b)$ and $P(k_b)$ the
frame-averaged radial log-spectra of member $k$ and of the observation, and by
$K_{\mathrm{spec}}$ the number of members entering the penalty, the
spatial component is
\begin{equation}
  \mathcal{L}_{\mathrm{spec}}^{\mathrm{spat}} =
    \frac{1}{K_{\mathrm{spec}} n_b} \sum_{k=1}^{K_{\mathrm{spec}}} \sum_{b=1}^{n_b}
    \bigl\lvert \widehat{P}^{(k)}(k_b) - P(k_b) \bigr\rvert ,
  \label{eq:spec_spat}
\end{equation}
averaged over lead times. Because forecast and observation are processed
identically, the normalisation of Eq.~\eqref{eq:psd2d} and the azimuthal
factor of Eq.~\eqref{eq:psd_rad} are common to both terms and cancel in the
log-space difference.

A purely spatial constraint can be satisfied by a temporally frozen texture,
whose spectrum is correct at every frame while its temporal evolution is not.
To prevent this degenerate solution, a second component compares temporal
power spectra: a one-dimensional Fourier transform is applied per pixel along
the lead-time axis after removal of the temporal mean, the resulting power is
averaged over pixels within each of the same $64 \times 64$ windows used for
the spatial term, and the window log-spectra are averaged over window
positions, yielding one temporal log-spectrum per ensemble member and lead
time, with the zero-frequency bin discarded. Denoting by $\overline{Q}^{(k)}(f)$
and $\overline{Q}(f)$ the frame-averaged temporal log-spectra of member $k$ and
of the observation at temporal frequency $f = 1, \dots, \lfloor m/2 \rfloor$
($m$ being the forecast length of Sect.~\ref{sec:architecture}), the temporal
component is
\begin{equation}
  \mathcal{L}_{\mathrm{spec}}^{\mathrm{temp}} =
    \frac{1}{K_{\mathrm{spec}} \lfloor m/2 \rfloor}
    \sum_{k=1}^{K_{\mathrm{spec}}} \sum_{f=1}^{\lfloor m/2 \rfloor}
    \bigl\lvert \overline{Q}^{(k)}(f) - \overline{Q}(f) \bigr\rvert.
  \label{eq:spec_temp}
\end{equation}
Both components enter the spectral penalty with equal weight,
$\mathcal{L}_{\mathrm{spec}} = \mathcal{L}_{\mathrm{spec}}^{\mathrm{spat}} +
\mathcal{L}_{\mathrm{spec}}^{\mathrm{temp}}$. To bound the memory cost of
the Fourier transforms, the penalty is evaluated on $K_{\mathrm{spec}} = 2$ randomly drawn
ensemble members at each training step, which leaves the estimator unbiased in
expectation.

The generator objective then becomes
\begin{equation}
  \mathcal{L}_{\text{total}} =
    \mathcal{L}_{\text{rec}}
    + w_D \lambda \mathcal{L}_{\text{adv}}
    + w_S \mathcal{L}_{\mathrm{spec}} ,
\end{equation}
where $w_S$ is a fixed scalar hyperparameter, set to $w_S = 0.02$. The
spectral term is deliberately kept outside $\mathcal{L}_{\text{rec}}$ when
computing the adaptive weight $\lambda$, so that adding it does not implicitly
rescale the adversarial contribution.

% TODO(check): the IRENE-GAN model evaluated in this paper is job33448 (resumed
% from job33387, epoch 28), whereas IRENE-GAN-RAPSD (job43955) was initialised
% from job37172, epoch 41. Verify whether job37172 is the continuation of
% job33448 -- in which case the sentence below is accurate -- or an independent
% branch, in which case the two adversarial models are siblings rather than
% parent and child and this paragraph must be reworded.
IRENE-GAN-RAPSD is not trained from scratch: it is initialised from the best
checkpoint of the IRENE-GAN training chain and fine-tuned with the spectral term
added to the generator objective, all other settings being unchanged and the
discriminator active from the first fine-tuning step. Its scores therefore
isolate the effect of adding the spectral penalty to an already adversarially
trained model, rather than the effect of optimising the two terms jointly from
the start. 
\subsection{Training setup}
\label{sec:training_setup}

The optimisation settings of the three configurations are summarised in
Table~\ref{tab:training}. All models are trained on $256 \times 256$ patches of
18 timesteps, with 6 input and 12 target frames, using Adam with a learning rate
of $10^{-4}$ on a single GPU, and the afCRPS is evaluated on $K = 10$ ensemble
members. In the adversarial configurations the generator and the discriminator
are updated alternately at each step by two separate optimisers sharing the same
learning rate, and the discriminator configuration is identical in IRENE-GAN and
IRENE-GAN-RAPSD, so that differences between the two can be attributed to the
spectral penalty. Training is stopped when the validation loss ceases to improve
and the checkpoint with the lowest validation loss is retained. 
% One difference should be noted: in IRENE the afCRPS is masked, i.e. evaluated only on the grid points where the radar composite provides valid data, whereas in the two adversarial configurations it is evaluated on all grid points of the patch, including the few residual invalid ones left after the filtering of Sect.~\ref{sec:dataset}.

\begin{table}[t]
  \caption{Optimisation and loss hyperparameters of the three IRENE
  configurations. Dashes denote entries that do not apply. The discriminator
  architecture, its parameter count, its warm-up schedule, and the
  RAPSD-specific hyperparameters ($w_S$, spectral window, $n_b$, members per
  spectral step) are configuration-specific implementation details reported in
  the text (Sect.~\ref{sec:train_gan} and Sect.~\ref{sec:train_gan_rapsd})
  rather than tabulated here.}
  \label{tab:training}
  \begin{tabular}{llll}
    \toprule
     & IRENE & \begin{tabular}[c]{@{}l@{}}IRENE\\ -GAN\end{tabular} & \begin{tabular}[c]{@{}l@{}}IRENE-GAN\\ -RAPSD\end{tabular} \\
    \midrule
    Optimiser & Adam & Adam & Adam \\
    Learning rate & $10^{-4}$ & $10^{-4}$ & $10^{-4}$ \\
    Batch size & 32 & 16 & 32 \\
    Training ensemble $K$ & 10 & 10 & 10 \\
    afCRPS fairness $\alpha$ & 0.95 & 0.95 & 0.95 \\
    $\lambda_{\mathrm{temp}}$ & 0.01 & 0.01 & 0.01 \\
    Discriminator width & -- & 16 & 16 \\
    Adversarial weight $w_D$ & -- & 0.04 & 0.04 \\
    \bottomrule
  \end{tabular}
\end{table}

\section{Evaluation metrics}

\label{sec:metrics}

Model performance is assessed using complementary deterministic and probabilistic
metrics. Deterministic performance is quantified using the Mean Absolute Error
(MAE) of the ensemble mean, while probabilistic skill is evaluated using the
Continuous Ranked Probability Score \citep[CRPS,][]{gneiting2007} and rank
histograms \citep{hamill2001}. In addition, we compute the radial power spectral
density (PSD) to compare the spatial sharpness of the forecasts with the ground
truth. All metrics are evaluated at each
forecast lead time $\tau \in \{1, \dots, T\}$ and averaged over the evaluation
period and spatial domain.

\subsection{Mean Absolute Error (MAE)}
\label{sec:metric_mae}

To evaluate the models trained with a pixel-wise
reconstruction objective, we use the Mean Absolute Error (MAE). Let
$\hat{y}_{i,\tau}$ denote the ensemble mean forecast at spatial grid point $i$
and lead time $\tau$, and let $y_{i,\tau}$ be the corresponding observed radar
rainfall. For a spatial domain of $N$ grid points, the spatial MAE at lead time
$\tau$ is defined as
\begin{equation}
  \mathrm{MAE}(\tau) = \frac{1}{N} \sum_{i=1}^{N} \bigl|\hat{y}_{i,\tau} - y_{i,\tau}\bigr|.
\end{equation}
MAE provides a straightforward and robust measure of grid-point accuracy by quantifying the average magnitude of absolute errors between predicted and observed precipitation. It offers an interpretable assessment of typical prediction errors across the spatial domain and lead times. However, as a purely point-wise metric, it does not explicitly capture spatial structure or displacement effects, and therefore does not directly reflect the realism or sharpness of predicted precipitation patterns.

\subsection{Continuous Ranked Probability Score}
\label{sec:metric_crps}

The Continuous Ranked Probability Score (CRPS) is a strictly proper scoring
rule that generalises the mean absolute error to probabilistic forecasts. Being a
strictly proper scoring rule, the CRPS is minimised in expectation if and only
if the CDF, $F$, equals the true data-generating distribution, simultaneously rewarding
calibration and sharpness.
The CRPS is evaluated pointwise at each grid point and lead time using the ensemble of predicted intensities, and then averaged over space and time.
The mean CRPS is obtained by averaging
Eq.~\eqref{eq:crps_ensemble} over all grid points, $i$, in the set of spatial points, $\Omega$, and all
evaluation samples in the test set:
\begin{equation}
  \overline{\text{CRPS}}(\tau) =
    \frac{1}{|\Omega| \cdot N_{\text{test}}}
    \sum_{n=1}^{N_{\text{test}}} \sum_{i \in \Omega}
    \widehat{\text{CRPS}}_K\!\left(\{y^{(k)}_{n,i,\tau}\}, y_{n,i,\tau}\right),
  \label{eq:mean_crps}
\end{equation}
where $y_{n,i,\tau}$ denotes the observed radar intensity at grid point $i$,
lead time $\tau$, and test sample $n$. A lower value of
$\overline{\text{CRPS}}$ indicates better probabilistic skill.
 
\subsection{Rank Histogram}
 
The rank histogram (also known as the Talagrand diagram) provides a
non-parametric diagnostic of ensemble calibration. For a scalar observation $y$
and a $K$-member ensemble $\{y^{(k)}\}_{k=1}^{K}$, the rank $r$ of $y$ within
the augmented sequence $(y^{(1)}, \dots, y^{(K)}, y)$ is computed as
\begin{equation}
  r = 1 + \sum_{k=1}^{K} \mathds{1}_{\{y^{(k)} < y\}}
        + \left\lfloor U \left(1 + \sum_{k=1}^{K} \mathds{1}_{\{y^{(k)} = y\}}\right)\right\rfloor,
  \label{eq:rank}
\end{equation}
so that $r \in \{1, \dots, K+1\}$. The first term counts the ensemble members
strictly below $y$, and $U$ is an independent random variable drawn uniformly
from $[0,1)$ for each grid point and evaluation sample. Denoting by
$t = \sum_{k=1}^{K} \mathds{1}_{\{y^{(k)} = y\}}$ the number of ensemble
members exactly tied with $y$, the floor $\left\lfloor U(1+t) \right\rfloor$
is uniformly distributed over the integers $\{0, 1, \dots, t\}$; the second
term of Eq.~\eqref{eq:rank} therefore resolves ties by drawing $r$ uniformly
among the $t+1$ ranks consistent with the block of members equal to $y$,
following standard practice for verifying variables with discretised or
bounded support \citep{hamill2001}. Ties of this kind arise systematically for radar-derived
precipitation, whose distribution has a point mass at zero: prior to rank
computation, observed and forecast intensities below the common detectability
threshold of $0.037$\,mm\,h$^{-1}$ are set to zero for every model, so that
the no-precipitation state is represented identically across observations and
all forecasting systems irrespective of their native minimum resolvable
intensity. Ranks are accumulated with the \texttt{pysteps} implementation
\citep{pulkkinen2019}. The rank histogram is the empirical frequency
distribution of $r$ over all valid (non-missing) grid points and evaluation
samples.
% TODO: verify citation key `hamill2001` against references_additions.bib
% (Hamill, T.M., 2001: Interpretation of Rank Histograms for Verifying
% Ensemble Forecasts)
 
For a perfectly calibrated ensemble, observations are statistically
indistinguishable from ensemble members, and the rank histogram should be 
uniform. Systematic deviations from uniformity diagnose specific
issue in the forecast:
\begin{itemize}
  \item a \emph{U-shaped} histogram indicates ensemble underdispersion (too
    narrow spread relative to the observations);
  \item a \emph{dome-shaped} histogram indicates overdispersion;
  \item a monotone slope indicates a systematic bias in the ensemble mean.
\end{itemize}
The rank histogram is computed both at each forecast lead
time $\tau$, to track the evolution of calibration as the forecast horizon
increases, and for the overall forecast.
 
To condense this lead-time-resolved diagnostic into a single summary curve, we
also report the root-mean-square deviation (RMSD) of the rank histogram from
uniformity,
\begin{equation}
  \begin{aligned}[t]
    \mathrm{RMSD}(\tau) &= \\
    &\sqrt{\frac{1}{K+1}\sum_{r=1}^{K+1}
      \left(p_r(\tau) - \frac{1}{K+1}\right)^2} ,
  \end{aligned}
  \label{eq:rmsd}
\end{equation}
where $p_r(\tau)$ is the empirical frequency of rank $r$ at lead time $\tau$
(the bar heights of the rank histogram). Larger values of $\mathrm{RMSD}(\tau)$
indicate a stronger departure from the ideal uniform histogram; this quantity
is shown as a function of lead time in the bottom-right panel of
Figure~\ref{fig:rkh}.

\subsection{Radial power spectral density}

The comparison of radial power spectral densities (PSDs) provides a description of how different models distribute variance across spatial scales. It reveals whether a model reproduces the observed large-scale and small-scale variability, or whether it exhibits systematic deficiencies such as excessive smoothing or spurious high-wavenumber noise. A model whose spectrum lies below the reference at large wavenumbers underrepresents fine-scale structure, whereas an excess of power at those scales indicates artificial small-scale variability. Agreement at low wavenumbers but divergence at high wavenumbers suggests that the model captures the dominant large-scale features while failing to represent convective structures.

As a verification diagnostic, the radial PSD of Eq.~\eqref{eq:psd_rad} is
evaluated on the full forecast domain, at each lead time and for each ensemble
member, rather than on the subdomains used for the training penalty of
Sect.~\ref{sec:train_gan_rapsd}. The diagnostic therefore resolves scales from
the domain scale to the Nyquist limit, and is applied identically to all
models, including those not trained with a spectral constraint.

Excess power is defined as the difference, in dB, between a model's radially averaged power spectral density and that of the observation at a given scale and lead time; positive values indicate that the model generates more power (i.e., more fine-scale variance) than observed, while negative values indicate a deficit.

\subsection{Benchmark models}

To comprehensively evaluate the predictive skill of the proposed IRENE configurations, we compare their performance against two established benchmark models. These baselines were selected to represent the two primary paradigms in modern precipitation nowcasting: traditional operational extrapolation and state-of-the-art deep learning. Specifically, we employ the Short-Term Ensemble Prediction
System (STEPS)~\citep{bowler2006} as the standard for probabilistic optical-flow methods, providing a robust baseline for short-term kinematic advection. As the deep learning benchmark, we use the Deep Generative Model of Rainfall (DGMR)~\citep{ravuri2021}, a well-established adversarial architecture for generative precipitation forecasting. Together, these models provide a comprehensive framework to assess the relative advantages of IRENE.

\subsubsection{STEPS}

As a classical probabilistic baseline, we use STEPS to generate ensemble nowcasts from the
same IT-DPC-SRI radar composite input. STEPS combines Lagrangian advection of
observed radar fields with stochastic perturbations designed to reproduce the
observed scale-dependent variability of precipitation fields. Operational
implementations at kilometre-scale resolution and 5-minute timestep---such as
those used by the UK Met Office and MeteoSwiss---provide a strong reference for
evaluating deep-learning-based nowcasting systems.

In our implementation, STEPS is run via the \texttt{pysteps} library
\citep{pulkkinen2019} with the following configuration. The motion field is
estimated from consecutive radar frames using the Lucas--Kanade optical-flow
algorithm~\citep{lucas1981}. The stochastic forecast is then generated over
$N = 24$ lead times (120\,min) using 6 cascade levels and a second-order
autoregressive (AR(2)) model for the temporal evolution of each level.
Stochastic noise is generated using a non-parametric method and perturbed
velocity fields are added following the scheme of \citet{bowler2006}.
Ensemble members are post-processed using CDF matching against the observed
precipitation distribution to correct systematic biases in the ensemble
marginal distribution. Precipitation below a threshold of
$5.0\,\text{dBZ}$ is masked using an incremental masking strategy.
The spatial resolution and timestep are set to match the IT-DPC-SRI composite
at \SI{1}{km\per pixel} and \SI{5}{min}, respectively. STEPS is initialised with the same 6 input frames
(\SI{30}{min}) as IRENE, and the number of ensemble
members $K$ is set to 10, equal to that used at inference by all models.

\subsubsection{DGMR}

The Deep Generative Model for Rainfall (\textsc{DGMR})~\citep{ravuri2021}
is a state-of-the-art deep learning baseline for probabilistic precipitation
nowcasting. Built around an architecture featuring context conditioning and a
convolutional gated recurrent unit (ConvGRU) sampler, DGMR is trained via a
purely adversarial framework. It employs both spatial and temporal
discriminators to enforce structural realism and meteorological consistency
across the generated forecast trajectories. Its ability to produce
probabilistically sharp and realistic precipitation fields has demonstrated
competitive skill against expert human forecasters in operational settings,
making it a natural benchmark for IRENE.

We used the pre-trained DGMR model provided with the official implementation \citep{deepmind_nowcasting}. The available pre-trained weights are trained on the sample split of the UK Nimrod 1 km radar dataset, and the framework operates on 24-timestep sequences, consistent with the original DGMR nowcasting setting of forecasts up to 90 minutes ahead.

Two consequences of this choice must be kept in mind when interpreting the
results. First, DGMR is applied here in a purely out-of-distribution setting: it
was trained on a different radar network, over a different orographic and
climatological regime, and no fine-tuning on the Italian composite was
performed. Its scores therefore quantify the transferability of a pre-trained
nowcasting model rather than the intrinsic skill of the DGMR architecture, and
they should not be read as an upper bound of what that architecture could
achieve if retrained on the IT-DPC-SRI archive. Second, its input and output lengths are those of the released model, namely 4
input frames (\SI{20}{min}, against \SI{30}{min} for IRENE and STEPS) and 18
forecast steps: all DGMR curves in Sect.~\ref{sec:results} therefore terminate
at \SI{90}{min} and no comparison is possible over the last \SI{30}{min} of the
IRENE and STEPS forecasts. As for the other models, 10 ensemble members are
generated.

A similar caveat applies to STEPS: the CDF matching against the observed
precipitation distribution and the incremental masking below
$5.0\,\text{dBZ}$ act as a statistical calibration of the marginal
distribution, which is expected to favour STEPS in the spectral and rank
diagnostics relative to the purely learned models, none of which receives an
equivalent post-processing.

% Table: summary of benchmark models
%\begin{table*}[t]
%  \centering
%  \caption{Summary of benchmark models used for comparison with IRENE.}
%  \label{tab:benchmarks}
%  \begin{tabular}{llp{.69\textwidth}}
%    \toprule
%    Model & Type & Brief description \\
%    \midrule
%    STEPS & Extrapolation & Lagrangian radar extrapolation with stochastic perturbations. \\
%    DGMR & GAN-based & Conditional GAN with recurrent convolutional generator and spatial and temporal discriminator. \\
%    \bottomrule
%  \end{tabular}
%\end{table*}

\section{Results}
\label{sec:results}

\begin{figure*}
  \centering
  \includegraphics{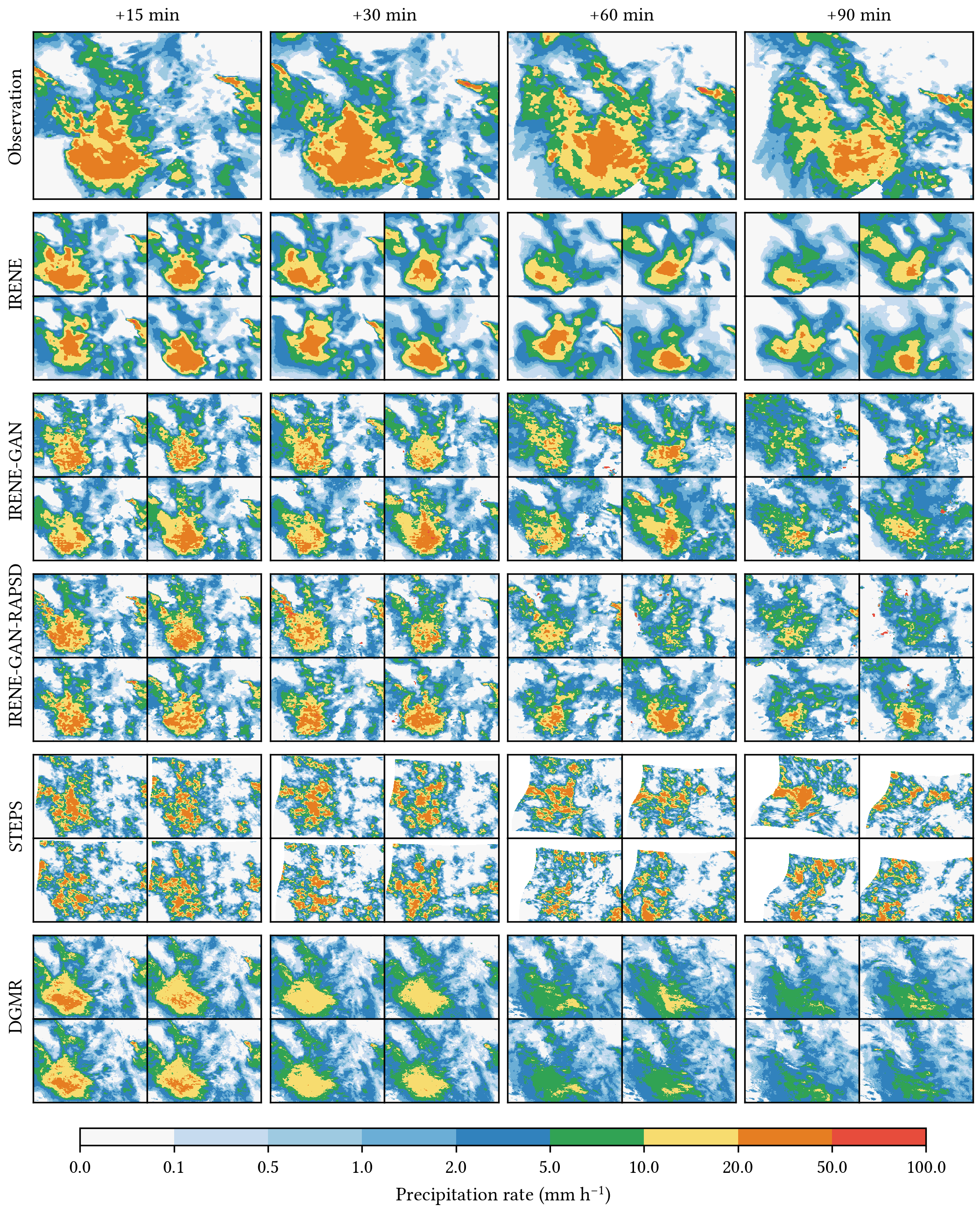}
  \caption{Comparison of a sample forecast from different models, for a
  $256 \times 256$\,km patch centred at approximately $42.5^{\circ}$N,
  $7.3^{\circ}$E over the Ligurian Sea and clipping the north-western
  Italian--French coastline (bounding box
  $41.33$--$43.75^{\circ}$N, $5.67$--$8.95^{\circ}$E), with the 6-frame input
  window starting 2025-11-02T18:10~UTC. Columns correspond to lead times of
  $+15$, $+30$, $+60$ and $+90$~min relative to the end of the input window.
  The top row shows the radar observation; each of the following rows shows,
  for one model, a $2 \times 2$ block with four ensemble members. The
  straight-edged blank margins in the STEPS panels at long lead times are the
  areas for which the Lagrangian extrapolation provides no source pixels.}
  \label{fig:forecast_comparison}
\end{figure*}

This section presents quantitative and qualitative results for IRENE and a
comparison with the benchmark models, evaluated using the metrics introduced
in Section~\ref{sec:metrics}. All metrics are computed exclusively on the test dataset and over
valid grid points, defined as those for which all models produce a finite
forecast value at a given lead time. Specifically, a common validity mask is
applied at each lead time $\tau$ before computing any metric, excluding grid
points where at least one model returns a missing value. This is particularly
relevant when comparing against \textsc{STEPS}, whose Lagrangian advection
scheme progressively advects precipitation fields,
introducing NaN-valued boundary regions that grow with lead time. Restricting
evaluation to the common valid domain ensures that all models are assessed
on an identical set of grid points, avoiding any artificial skill differences.

Note that the models are trained on a forecast horizon of \SI{60}{min} but, being fully
recurrent, they can be rolled out for an arbitrary number of steps at inference.
All results in Sect.~\ref{sec:results} are produced with a \SI{120}{min}
rollout, i.e. twice the training horizon, so that the second hour of the
forecast is an extrapolation beyond the regime seen during training.

A similar argument applies to the spatial extent of the forecast. Training and
verification in this paper use $256\times256$ patches (Sect.~\ref{sec:dataset}),
but the encoder--forecaster contains no operation with a fixed spatial size:
every layer is either a $3\times3$ convolution, a ConvGRU cell, or a
PixelShuffle/PixelUnshuffle operation, each of which acts identically
regardless of input height and width, up to padding to a multiple of
$2^{S} = 32$ pixels. The same trained weights can therefore be applied, in a
single forward pass and without retraining, to the full national grid targeted
by the operational IT4LIA deployment, rather than requiring the domain to be
split into tiles that are forecast separately and stitched back together; no
patch-boundary artefacts of the kind produced by overlap-and-stitch inference
are consequently expected.

\subsection{Case studies}

As a first comparison, we present a case study of a high-impact precipitation
event over the Ligurian Sea, selected among the test-period cases for its high
total precipitation. The synoptic timing and location, an intense system over
the Ligurian Sea in early November, are typical of a so-called Genoa low: a
Mediterranean cyclone that develops in the lee of the Alps and is a recurrent
driver of severe autumn precipitation along the north-western Italian coast.
Figure~\ref{fig:forecast_comparison} shows the observed
sequence together with four ensemble members of each model at $+15$, $+30$,
$+60$ and $+90$ min. The observed field consists of a large, well-organised
cluster with an extended core exceeding
$20$\,mm\,h$^{-1}$, embedded in a broad region of light to moderate
precipitation; the cluster propagates north-eastwards while the intense core
persists throughout the two hours.

IRENE places the main precipitation areas in broadly the correct positions, and
its members reproduce the intense core with the correct order of magnitude up to
$+30$ min. The fields are, however, visibly smoother than the observations at
all lead times: the core appears as a single compact blob without internal
structure, and the surrounding light precipitation is rendered as broad
homogeneous areas, with part of the weak-rain field disappearing altogether
beyond $+60$ min. This is the expected signature of a purely probabilistic
point-wise objective, which is minimised by hedging towards the conditional
median of the small-scale variability, and it corresponds to the spectral
deficit quantified in Sect.~\ref{sec:results_psd}.

IRENE-GAN and IRENE-GAN-RAPSD produce visually much more realistic fields: the
texture of the stratiform region is restored, and the light-precipitation areas
that IRENE erases are retained until the end of the forecast. The intense core,
however, is no longer represented as a coherent area but as a dense
speckle of high-intensity pixels embedded in a moderate-intensity
background, i.e. the fine-scale variance is reinstated with an incorrect spatial
organisation. The two adversarial configurations are nearly indistinguishable at
$+15$ and $+30$ min; at $+60$ and $+90$ min IRENE-GAN-RAPSD weakens the core
more than IRENE-GAN, retaining fewer pixels above $20$\,mm\,h$^{-1}$, and
occasionally produces isolated pixels above $50$\,mm\,h$^{-1}$ that have no
counterpart in the observations.

STEPS advects the field along a trajectory very similar to that of the learned
models, indicating that the motion inferred by IRENE is consistent with the
optical-flow estimate, and it preserves both the observed texture and the
observed intensity distribution, since the Lagrangian extrapolation transports
the initial field rather than regenerating it. Two artefacts are nevertheless
apparent at long lead times: the precipitation area becomes bounded by straight
edges, where the advected domain no longer provides source pixels, and the
intense cores are fragmented into small speckles distributed over an area larger
than the observed one.

The pre-trained DGMR reproduces the observed field convincingly during the first
$30$ min, including the position and the amplitude of the core. From $+60$ min
onwards, however, the intensities collapse: the core disappears and the whole
field is reduced to light and moderate precipitation, while the four members
remain very similar to one another. This joint loss of intensity and of spread
is the qualitative counterpart of the low MAE, high CRPS and strongly U-shaped
rank histogram discussed below.

\subsection{Deterministic performance}

We first compare the ensemble mean of the three IRENE configurations against the
ensemble mean of STEPS and DGMR.

\begin{figure}[t]
  \centering
  \includegraphics{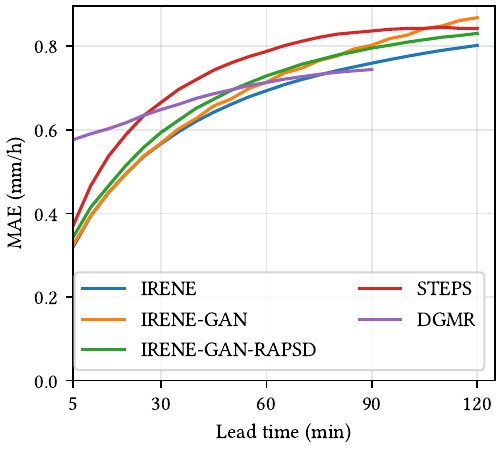}
  \caption{MAE of the ensemble mean of IRENE, IRENE-GAN, IRENE-GAN-RAPSD, STEPS
  and DGMR as a function of lead time. The DGMR curve terminates at
  \SI{90}{min}, the maximum horizon of the pre-trained model.}
  \label{fig:mae}
\end{figure}

Figure \ref{fig:mae} shows the MAE as a function of lead time for the different
models; lower values indicate better forecast skill. The three IRENE
configurations achieve the lowest MAE up to about \SI{85}{min}, with IRENE
attaining the best score among the IRENE configurations at all lead times, followed by IRENE-GAN-RAPSD and
IRENE-GAN. DGMR is by far the least accurate model over the first \SI{25}{min}
($0.58$ against $0.32$--$0.37$\,mm\,h$^{-1}$ at \SI{5}{min}), which is
consistent with limited out-of-domain generalisation from the UK training
dataset to the Italian study area. Its error, however, grows only marginally
with lead time, so that DGMR overtakes STEPS at around \SI{27}{min} and, beyond
approximately \SI{85}{min}, attains the lowest MAE of all models within its
\SI{90}{min} horizon. This behaviour is attributable to the excessive damping of
precipitation fields after approximately \SI{40}{min} in all ensemble members
(see, e.g., the last row of Fig. \ref{fig:forecast_comparison}): smoother
forecasts, and in particular those that suppress high precipitation intensities,
are systematically favoured by the MAE, which is why this metric alone is not a
meaningful ranking criterion for probabilistic nowcasts.

The same mechanism explains the ordering among the IRENE configurations. The
MAE of IRENE-GAN increases almost linearly beyond \SI{60}{min}, reflecting the
model's tendency to preserve or generate localised intense precipitation rather
than damp it, and it is the largest of all models at \SI{120}{min}
($0.87$\,mm\,h$^{-1}$). IRENE-GAN-RAPSD follows a similar trajectory but remains
consistently below IRENE-GAN beyond \SI{60}{min} ($0.83$\,mm\,h$^{-1}$ at
\SI{120}{min}), consistent with the RAPSD penalty moderating, without
suppressing, the generation of localised precipitation extremes. The MAE of
STEPS plateaus after about \SI{90}{min} and, from approximately \SI{110}{min}
onwards, falls below that of IRENE-GAN, while it remains above IRENE-GAN-RAPSD
and IRENE over the whole forecast horizon. The persistent difference between
IRENE and the two adversarial configurations is therefore not evidence of a loss
of forecast quality, but the expected penalty incurred by sharper fields under a
point-wise metric; the spectral diagnostics of Sect.~\ref{sec:results_psd} quantify
the corresponding gain in spatial realism.

\subsection{Probabilistic skill and calibration}

\begin{figure}
  \centering
  \includegraphics{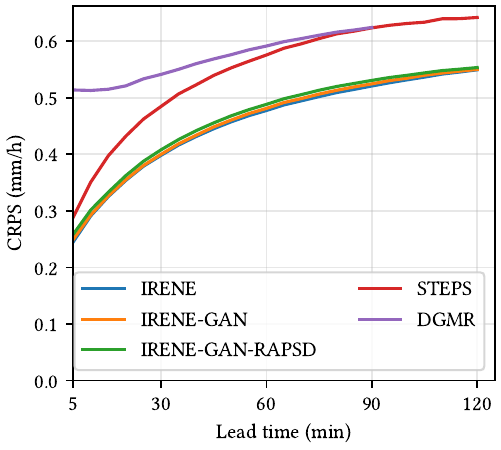}
  \caption{CRPS of IRENE, IRENE-GAN, IRENE-GAN-RAPSD, STEPS and DGMR as a
  function of lead time. The DGMR curve terminates at \SI{90}{min}, the maximum
  horizon of the pre-trained model.}
  \label{fig:crps}
\end{figure}

\begin{figure*}
  \centering
  \includegraphics{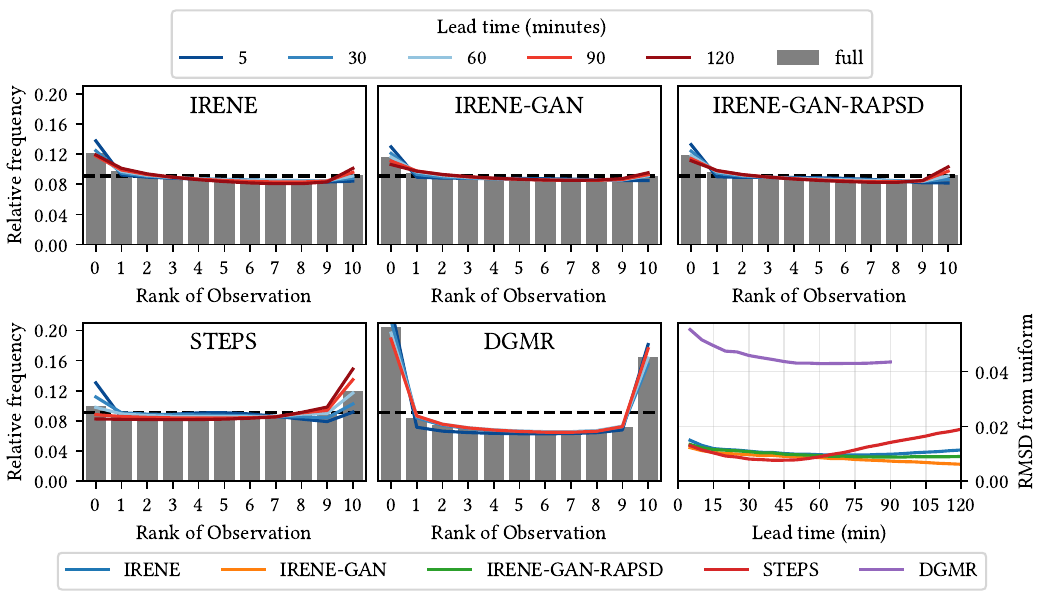}
  \caption{Rank histograms of IRENE, IRENE-GAN, IRENE-GAN-RAPSD, STEPS and DGMR
  for selected lead times (coloured lines) and for the full forecast (grey
  columns); the dashed line marks the uniform frequency $1/(K+1)$ expected for a
  calibrated ensemble. The bottom-right panel shows the root-mean-square
  deviation of the rank histogram from uniformity as a function of lead time,
  which summarises the departure from calibration of each model.}
  \label{fig:rkh}
\end{figure*}

Figure \ref{fig:crps} shows the CRPS as a function of lead time, which
complements the MAE by jointly rewarding accuracy and a correctly dispersed
ensemble. All three IRENE configurations outperform both benchmarks at every
lead time and achieve very similar CRPS to one another throughout the forecast
horizon: IRENE, trained directly with the afCRPS loss, is marginally the best,
while IRENE-GAN and IRENE-GAN-RAPSD remain within $0.005$\,mm\,h$^{-1}$ of it at
\SI{120}{min} ($0.549$ and $0.553$ against $0.548$\,mm\,h$^{-1}$). The
adversarial and spectral terms therefore introduce, at most, a marginal
degradation of probabilistic skill while substantially improving the spectral
properties of the fields (Sect.~\ref{sec:results_psd}). STEPS is
markedly worse at all lead times and, like the MAE, its CRPS flattens beyond
approximately \SI{90}{min}, reaching $0.64$\,mm\,h$^{-1}$ at \SI{120}{min}, some
$17\%$ above the IRENE configurations; unlike the MAE, however, it never
approaches them. DGMR has the worst CRPS over its entire \SI{90}{min} horizon,
converging towards the STEPS values only at the end of that horizon. The
contrast with the MAE ranking of DGMR shows that its apparently competitive
deterministic error at long lead times results from smoothing rather than from
skill, and disappears once the ensemble distribution is scored.

Forecast calibration is assessed with the rank histograms of Fig.
\ref{fig:rkh}. The three IRENE configurations are close to uniform at all lead
times, with a moderate excess of the lowest rank at short lead times
(frequency $\approx 0.12$ against the expected $1/11 \approx 0.09$) and a
comparable excess of the highest rank at \SI{90}{min} and \SI{120}{min}; the
corresponding deviation from uniformity remains below $0.015$ throughout
(bottom-right panel of Fig. \ref{fig:rkh}), and is smallest for IRENE-GAN at
long lead times. The two benchmarks behave differently. STEPS is nearly uniform
in the interior of the histogram but develops a pronounced excess of the highest
rank that grows with lead time (up to $\approx 0.15$ at \SI{120}{min}),
indicating a systematic underprediction of the most intense rain rates at long
horizons; its deviation from uniformity accordingly increases beyond
\SI{60}{min}, whereas that of the IRENE configurations does not. DGMR shows a
strongly U-shaped histogram at every lead time, with both extreme ranks close to
$0.17$--$0.20$ and interior frequencies near $0.065$. Following the
classification of Sect.~\ref{sec:metrics}, this is the signature of a markedly
underdispersive ensemble rather than of a conditional bias, in agreement with
the limited spread visible among its members in Fig.
\ref{fig:forecast_comparison}, and it yields a deviation from uniformity three
to five times larger than that of any other model.

\subsection{Power spectral density}
\label{sec:results_psd}

\begin{figure*}
  \centering
  \includegraphics{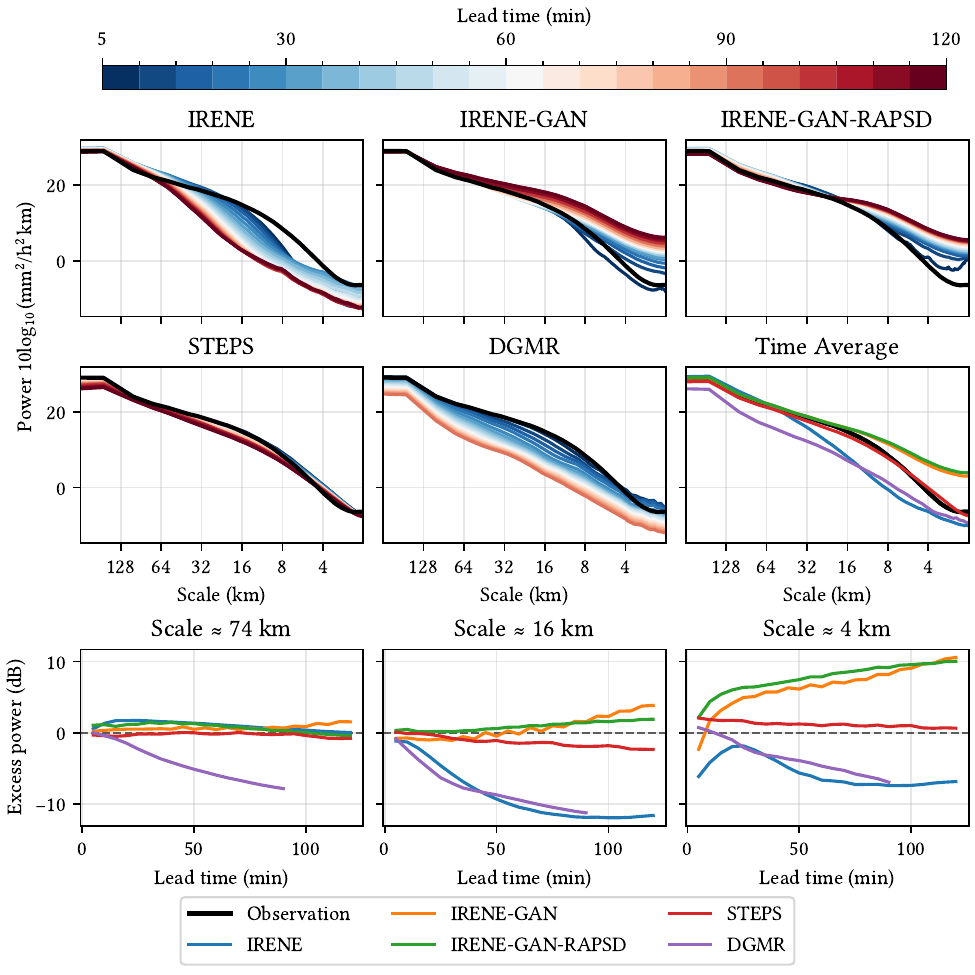}
  \caption{Top two rows: radial power spectral density as a function of spatial
  scale, colour-coded by forecast lead time, for IRENE, IRENE-GAN,
  IRENE-GAN-RAPSD, STEPS, and DGMR, plus the lead-time-averaged spectrum of all
  models (rightmost panel, second row). The observation is shown as the thick
  black line. Bottom row: excess power relative to the observation (in dB) as a
  function of lead time, evaluated at three representative scales
  (\(\approx\)74, \(\approx\)16, and \(\approx\)4 km). The DGMR curves
  terminate at \SI{90}{min}, the maximum horizon of the pre-trained model.}
  \label{fig:rpsd_leadtime}
\end{figure*}

Figure~\ref{fig:rpsd_leadtime} shows the evolution of the radial power spectral density with lead time for the five forecasting systems, together with the lead-time-averaged spectrum and the excess power relative to the observation at three representative scales. STEPS, IRENE, and DGMR generally exhibit a decrease in power as lead time increases. For IRENE and DGMR, this behaviour is consistent with a progressive damping of precipitation intensity, and in the case of IRENE it is also expected from training with afCRPS and the temporal consistency penalty, which tend to smooth the field and progressively suppress power, particularly at smaller scales. By contrast, IRENE-GAN preserves substantially more fine-scale power and tends to increase power with lead time, especially at the smallest scales, indicating an excess of spatial detail.

IRENE-GAN-RAPSD, which augments the adversarial loss with an explicit penalty on the radially averaged power spectral density, improves agreement with the observed spectrum relative to IRENE-GAN, but the correction is scale-dependent. At the coarsest scale (\(\approx\)74 km), both GAN variants remain close to neutral throughout the forecast, within roughly ±2 dB of the observation. At the intermediate scale (\(\approx\)16 km), IRENE-GAN and IRENE-GAN-RAPSD track closely up to \(\approx\)80 min; beyond this, IRENE-GAN's excess power grows more steeply, reaching a higher value than IRENE-GAN-RAPSD by 120 min, indicating that the RAPSD penalty limits the accumulation of excess power specifically at longer lead times. At the finest resolvable scale (\(\approx\)4 km), however, this control is absent, and is in fact slightly counterproductive: both adversarial configurations reach approximately $+10$\,dB at 120 min, but IRENE-GAN-RAPSD starts from a larger excess at 5 min ($\approx +4$\,dB, against $\approx -2$\,dB for IRENE-GAN) and remains above IRENE-GAN for most of the forecast, the two curves converging only in the last 20 min. This indicates that the RAPSD penalty constrains the growth of excess power at coarse and intermediate scales but not at the smallest ones, where both adversarially trained models over-generate fine-scale detail at long lead times. A plausible explanation is that the training penalty is evaluated on $64\times 64$ windows and is therefore blind to the scales below its finest radial bin, so that the excess is displaced towards, rather than removed from, the highest wavenumbers.

Comparing the models with the observations, STEPS provides the closest overall agreement at short lead times and shows the best match when the spectra are averaged over lead times, but its skill degrades progressively at longer lead times. This degradation is most pronounced at the intermediate scale (\(\approx\)16 km), where excess power falls to approximately -2 to -3 dB by 120 min, while remaining closer to neutral at \(\approx\)74 km and \(\approx\)4 km. DGMR remains systematically below the observed power spectrum at nearly all scales and lead times, with its deficit growing fastest and largest among all models, reaching roughly -8 to -12 dB at \(\approx\)74 km and \(\approx\)16 km within the available forecast window ($\approx$90 min), reflecting an overall lack of variability and limited transferability outside its training distribution. IRENE begins to lose power already at scales between \(\approx\)16 and \(\approx\)74 km, and this loss is most pronounced at the intermediate scale (\(\approx\)16 km) at long lead times (approximately -12 dB by 120 min), while remaining comparatively closer to neutral at \(\approx\)74 km, consistent with the smoothing behaviour discussed above.

\section{Conclusions}

We have introduced IRENE, a multi-scale ConvGRU model for probabilistic radar
precipitation nowcasting over Italy, designed for operational deployment within
the IT4LIA AI Factory. The model is trained on the national DPC radar composite,
after selecting relevant precipitation events using an importance-sampling
scheme. Three configurations were trained and evaluated: a probabilistic
configuration based on the afCRPS loss (IRENE), an adversarial configuration
(IRENE-GAN), and a spectrally constrained adversarial configuration
(IRENE-GAN-RAPSD).

Our evaluation shows that all three configurations provide skillful
probabilistic forecasts, outperforming both the optical-flow benchmark (STEPS)
and the pre-trained deep learning baseline (DGMR) in terms of the Continuous
Ranked Probability Score at every lead time up to \SI{120}{min}, and yielding
rank histograms much closer to uniformity. STEPS reproduces the short-term
advection and the spatial structure of the precipitation events accurately, and
provides the best spectral agreement at short lead times, but its probabilistic
skill degrades faster and it increasingly underpredicts the highest rain rates
beyond \SI{60}{min}. DGMR, applied without fine-tuning outside its training
domain, damps the precipitation field after about \SI{40}{min}; this yields the
lowest ensemble-mean MAE at the end of its \SI{90}{min} horizon, but the worst
CRPS and a markedly underdispersive ensemble, illustrating both the difficulty
of transferring nowcasting models across climatological regimes and the fact
that point-wise errors alone are misleading for probabilistic systems.

The spectral analysis quantifies the trade-off between the configurations.
IRENE progressively loses variance at scales below \(\approx\)74 km, reaching a
deficit of about $-12$\,dB at \(\approx\)16 km by \SI{120}{min}, as expected
from a point-wise probabilistic loss. Adversarial training removes this deficit
but overshoots at long lead times, and the explicit RAPSD penalty only partly
corrects the overshoot: it reduces the excess at intermediate scales beyond
\SI{80}{min}, but is slightly counterproductive at the finest resolved scale
(\(\approx\)4 km), where IRENE-GAN-RAPSD's excess power exceeds that of
IRENE-GAN for most of the forecast. Reducing the
excess of fine-scale power at the smallest scales, for instance by evaluating
the spectral penalty on larger windows or by weighting the radial bins, is
therefore the most immediate line of improvement.

This study has three main limitations. First, the comparison with DGMR is not a
comparison of architectures, since the pre-trained model is used out of
distribution and its horizon is limited to \SI{90}{min}; a version retrained on
the Italian archive would be required for a like-for-like assessment. Second,
verification is restricted to point-wise and spectral scores; the skill on
intense, hydrologically relevant rain-rate thresholds and its scale dependence
remain to be quantified with categorical and neighbourhood scores. Third, the evaluation is performed on
instantaneous rain rates only; the ability of the models to reproduce
accumulated precipitation, which is what hydrological applications require, has
not been assessed here.

Future work will focus on refining both the architecture and the training
strategy, with particular attention to calibrating the GAN and spectral losses
so as to reduce excessive fine-scale structure in the precipitation field. In
addition, we plan to extend the model by incorporating additional predictors,
such as satellite and NWP fields, to verify the forecasts against rain-gauge
accumulations, and to investigate its integration within downstream hydrological
and impact-based forecasting systems.

\bibliographystyle{copernicus}
\bibliography{references,extra_refs}

\section{Code and Data Availability}

The IT-DPC-SRI ARCO datacube used for training and evaluation is described in
\citet{franch2026} and is available at \url{https://doi.org/10.5281/zenodo.18637608}. The updated repository with the IRENE source code, the training configurations of the three model variants, and the evaluation scripts will be made available soon under the  BSD 2-Clause License. The
trained model weights are archived at \url{https://huggingface.co/it4lia/irene/tree/main/scripts}. The
STEPS baseline was produced with the open-source \texttt{pysteps} library
\citep{pulkkinen2019}, and the DGMR baseline with the publicly released
pre-trained model \citep{deepmind_nowcasting}.

\section{Author Contributions}
Alessandro Camilletti and Gabriele Franch developed the software infrastructure and conducted the experiments. Elena Tomasi and Alessandro Camilletti developed the evaluation framework. Alessandro Camilletti analyzed the results. Alessandro Camilletti wrote the original draft of the manuscript. Marco Cristoforetti supervised the work and managed the project. All authors discussed the results and contributed to the final manuscript.

\section{Competing Interests}
The authors declare no competing interests.

\section{Acknowledgements}
This work was funded by the European Union through the EuroHPC Joint Undertaking under grant agreement No. 101234224 (IT4LIA — Italy for Artificial Intelligence), within the HORIZON-JU-EUROHPC-2025-AI-01-IBA-01 programme.

\end{document}